\documentclass[10pt,twocolumn,letterpaper]{article}

\usepackage{cvpr}
\usepackage{times}
\usepackage{epsfig}
\usepackage{graphicx}
\usepackage{amsmath}
\usepackage{amssymb}
\usepackage{subfigure}
\usepackage{multirow}

\usepackage[breaklinks=true,bookmarks=false]{hyperref}

\cvprfinalcopy 

\def\cvprPaperID{****} 

\begin{document}

\title{Selective Feature Connection Mechanism:\\ Concatenating Multi-layer CNN Features with a Feature Selector}

\author{Chen Du$^{1,2}$, Chunheng Wang$^{1}$,Yanna Wang$^{1}$, Cunzhao Shi$^{1}$, Baihua Xiao$^{1}$\\
$^{1}$Institute of Automation, Chinese Academy of Sciences(CASIA)  \\   $^{2}$University of Chinese Academy of Sciences(UCAS)\\
{\tt\small \{duchen2016, chunheng.wang,wangyanna2013, cunzhao.shi, baihua.xiao\}@ia.ac.cn}
}

\maketitle

\begin{abstract}
   Different layers of deep convolutional neural networks(CNNs) can encode different-level information. High-layer features always contain more semantic information, and low-layer features contain more detail information. However, low-layer features suffer from the background clutter and semantic ambiguity. During visual recognition, the feature combination of the low-layer and high-level features plays an important role in context modulation. If directly combining the high-layer and low-layer features, the background clutter and semantic ambiguity may be caused due to the introduction of detailed information. In this paper, we propose a general network architecture to concatenate CNN features of different layers in a simple and effective way, called Selective Feature Connection Mechanism (SFCM). Low-level features are selectively linked to high-level features with a feature selector which is generated by high-level features. The proposed connection mechanism can effectively overcome the above-mentioned drawbacks. We demonstrate the effectiveness, superiority, and universal applicability of this method on multiple challenging computer vision tasks, including image classification, scene text detection, and image-to-image translation.

\end{abstract}

\section{Introduction}

Deep Convolutional Neural Networks (CNNs) have achieved great success on a variety computer vision tasks, such as image classification \cite{krizhevsky2012imagenet}, semantic segmentation \cite{long2015fully,he2017mask}, and object detection \cite{girshick2015fast,ren2015faster,liu2016ssd,redmon2016you,dai2016r}.
In order to understand why the CNNs frameworks perform so well, Zeiler \emph {et al.}\cite{zeiler2014visualizing} introduce a novel visualization technique that gives insight into the function of intermediate feature layers and the operation of the classifier. Actually, feature maps of different layers extract information of different levels from input image. The low-layer features tend to response the patches with similar simple patterns and with more ambiguity. The feature maps of higher layers care more about semantic information but less detail information about an image, since higher layers are closer to the last layer with category labels.
\begin{figure}[!t]
\centering
\includegraphics[ width=8cm]{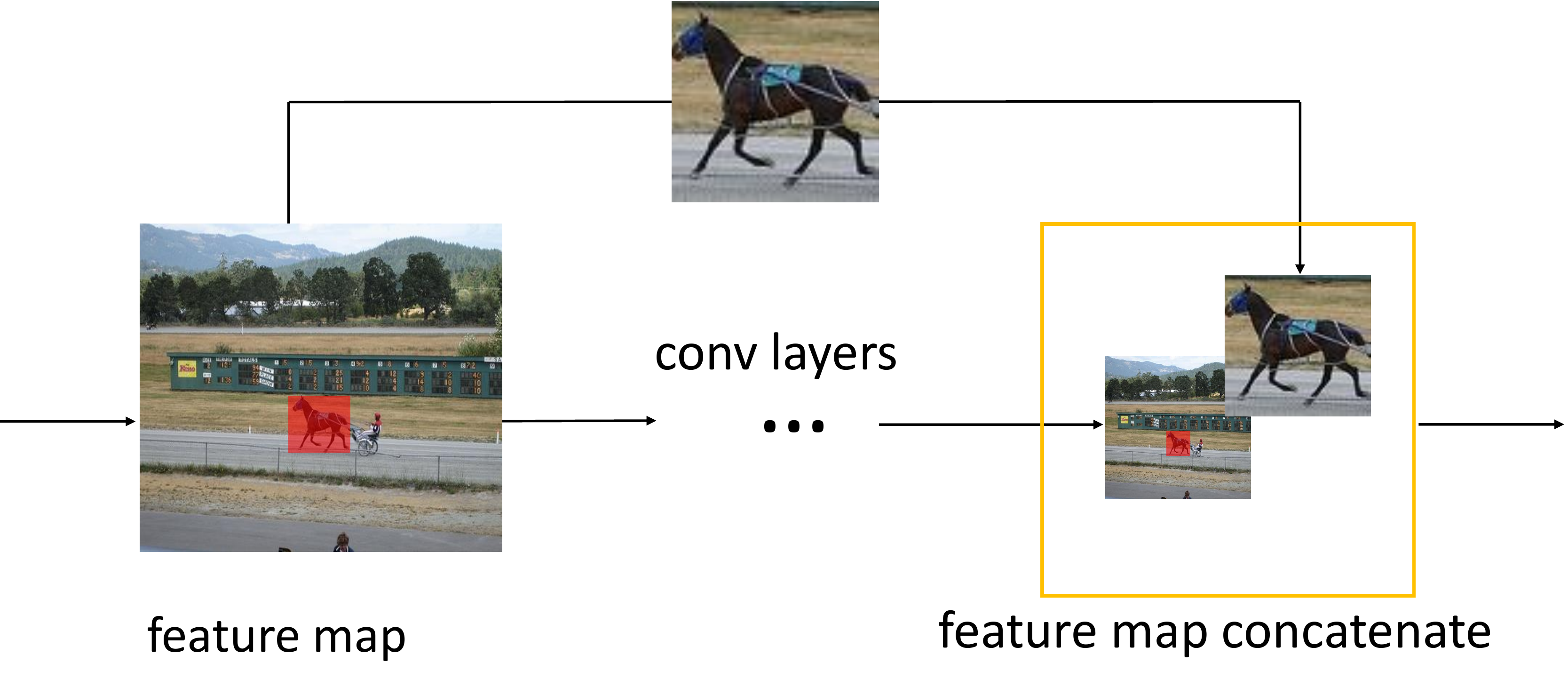}
\caption{During visual recognition, the feature combination of the low-layer and high-level features plays an important role in context modulation.}
\label{Fig1}
\end{figure}
Combining high-level information with low-level information effectively can improve the performance of CNNs in many computer vision tasks. The U-Net \cite{ronneberger2015u}, which combines the low-layer features and high-level features directly, works with very few training images and yields more precise segmentation. The EAST \cite{zhou2017east} adopt the idea from U-shape, merging feature maps gradually, to detect word regions of which the sizes vary tremendously. For many image translation problems, there is a great deal of low-level information shared between the input and output, and it would be desirable to shuttle this information directly across the net. Pix2pix \cite{isola2017image} is proposed to add skip connections for the generator to circumvent the information bottleneck. DenseNets \cite{huang2017densely} improve the flow of information and gradients throughout the network, which can also be seen as combining the low-layer features and high-level features. The FPN(Feature Pyramid Network) \cite{lin2017feature} develops a top-down architecture with lateral connections to build high-level semantic feature maps at all scales. This architecture shows significant improvement as a generic feature extractor in several applications.
\begin{figure*}[!t]
\centering
\includegraphics[width=16.5cm]{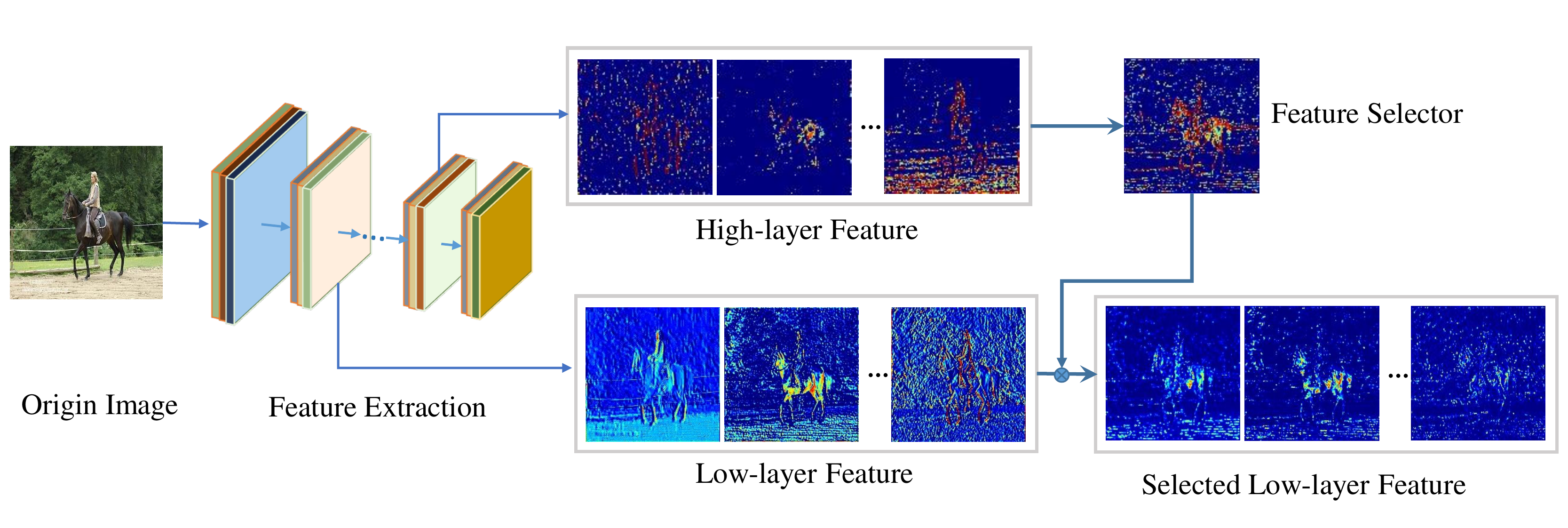}
\caption{ Different layers of deep convolutional neural networks(CNN) can encode different-level information. High-layer features always contain more semantic information, and low-layer features contain more detail information. In this paper, low-layer features are selected for further use by a feature selector which is generated by high-level features.}
\label{Fig2}
\end{figure*}

As shown in Fig.\ref{Fig1}, during visual recognition, the feature combination of the low-layer and high-level features plays an important role in context modulation. When observing a horse in a big image, we will pay more attention to the area that represents the detail of the horse. And in the human visual recognition system, high-level features play an important role to generate an attention map for selectively combining the low-layer features. But all these features combining methods \cite{ronneberger2015u,huang2017densely,lin2017feature} just combine the low-layer features and high-level features directly, which is not in accordance with the biological fact.

In this paper, we propose a general network architecture to concatenate CNN features of different layers in a simple and effective way, called \textbf{S}elective \textbf{F}eature \textbf{C}onnection \textbf{M}echanism (\textbf{SFCM}). Inspired by the human visual recognition mechanism, low-level features are selectively linked to high-level features with a feature selector which is generated by high-level features. Fig.\ref{Fig2} shows an example of low-layer features selected for further use by a feature selector which is generated by high-level features. We demonstrate the effectiveness, superiority, and universal applicability of this method on many challenging computer vision tasks, such as image classification, scene text detection, and image-to-image translation.

Our main contributions are as follows:
\begin{itemize}
\item We propose the SFCM, a general network architecture to concatenate CNN features of different layers in a simple and effective way. Different layers of deep ConvNet can encode different-level information. High-layer features always contain more semantic information, and low-layer features contain more detail information. However, low-layer features suffer from the background clutter and semantic ambiguity. The SFCM can be used in many existing frameworks to achieve better performance as it can combine high-resolution maps with low-level features without harm the semantic representation capacity of high-layer features.
\item We give two connection modes (direct connection and residual connection) to implement our SFCM. Both modes are effective and light weights. They add a small number of parameters and computation for the feature selector learning. They can be embedded into any existing deep CNNs based methods and the feature selector can be easily trained end-to-end with standard backpropagation.
\item The proposed SFCM was tested on many challenging computer vision tasks, such as image classification, scene text detection, and image-to-image translation. Experimental results show that our SFCM can be used in many existing frameworks to achieve better performance.
\end{itemize}

\section{Related Work}

\noindent\textbf{Network Architecture for improving NN performance.}
There are many notable network architecture innovations which have yielded competitive results. The Network in Network\cite{lin2013network} structure includes micro multi-layer perceptrons into the filters of convolutional layers to extract
more complicated features. Spatial Transformer Networks \cite{jaderberg2015spatial} is proposed to give neural networks the ability to actively spatially transform feature maps. \cite{dai2017deformable}introduce deformable convolution and deformable RoI pooling to enhance the transformation modeling capability of CNNs. In \cite{wang2016deeply}, Deeply-Fused Nets (DFNs) were proposed to improve information flow by combining intermediate layers of different base networks.

\noindent \textbf{Methods using multiple layers.} A number of recent approaches improve detection and segmentation by using different layers in a ConvNet. The U-Net \cite{ronneberger2015u}, which combines the low-layer features and high-level features directly, works with very few training images and yields more precise segmentation. FCN \cite{long2015fully} sums partial scores for each category over multiple scales to compute semantic segmentation. HyperNet \cite{kong2016hypernet}, ParseNet \cite{liu2015parsenet}, and ION \cite{bell2016inside} concatenate features of multiple layers before computing predictions, which is equivalent to summing transformed features. SSD \cite{liu2016ssd} and MS-CNN \cite{cai2016unified} predict objects at multiple layers of the feature hierarchy without combining features or scores. The FPN(Feature Pyramid Network) \cite{lin2017feature} develops a top-down architecture with lateral connections to build high-level semantic feature maps at all scales. But all these features combining methods just combine or use the low-layer features and high-level features directly, they have not considered the selectivity of high-level features to low-level features when the low-layer features are combined to the high-layer features.

\noindent \textbf{Attention mechanism.} Recently, attention mechanisms have been widely applied in different areas \cite{gregor2015draw,xu2015show,yang2016stacked}. Similar to human visual processing, attention-based models tend to selectively focus on the meaningful object. Vaswani et.al. \cite{vaswani2017attention} propose the self-attention method for machine translation. A self-attention module computes the response at a position in a sequence by attending to all positions and taking their weighted average in an embedding space. Based on the recent work of soft attention \cite{chen2016attention,jaderberg2015spatial}, the Residual Attention Network \cite{wang2017residual} is composed of multiple attention modules which generate attention-aware features. The attention aware features from different modules change adaptively as layers going deeper.

\begin{figure}[!t]
\centering
\includegraphics[width=8.5cm]{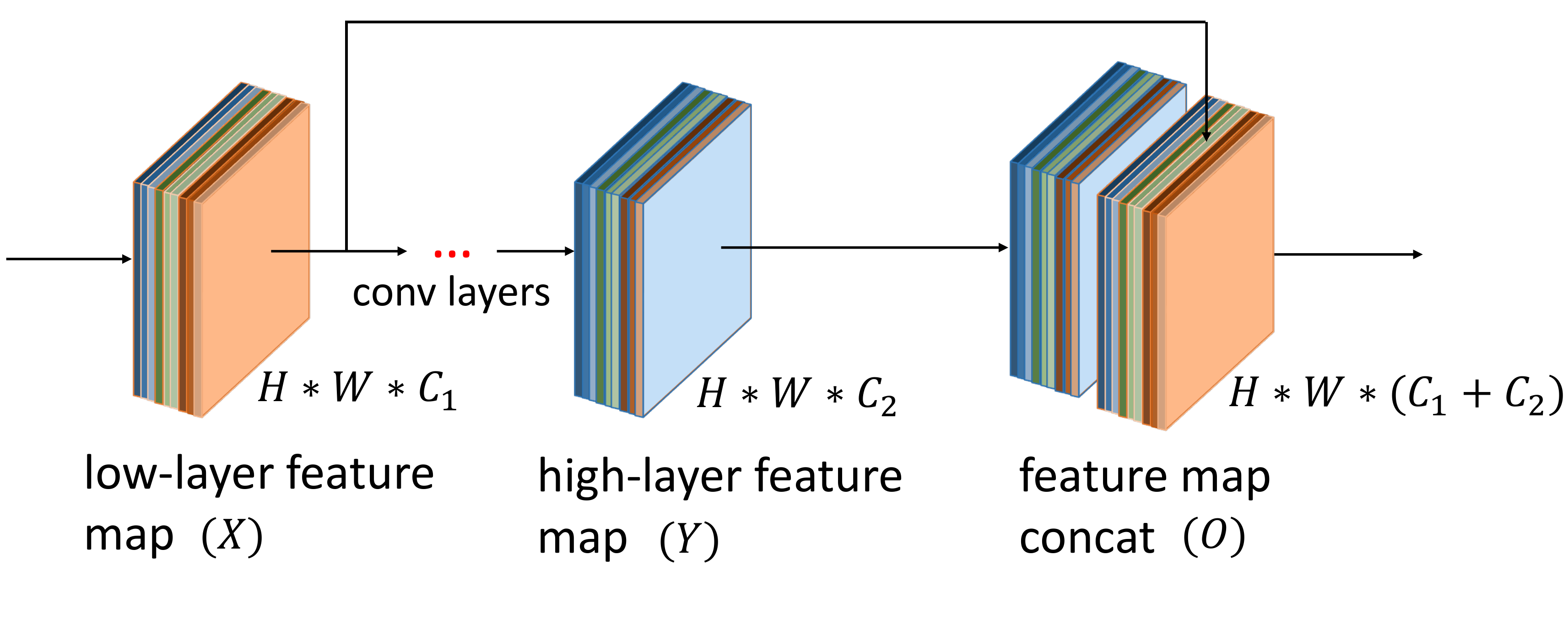}
\caption{Low-layer and high-layer features in CNN are combined directly.}
\label{Fig3}
\end{figure}

\begin{figure}[!t]
\centering
\subfigure[]{\includegraphics[scale=0.25]{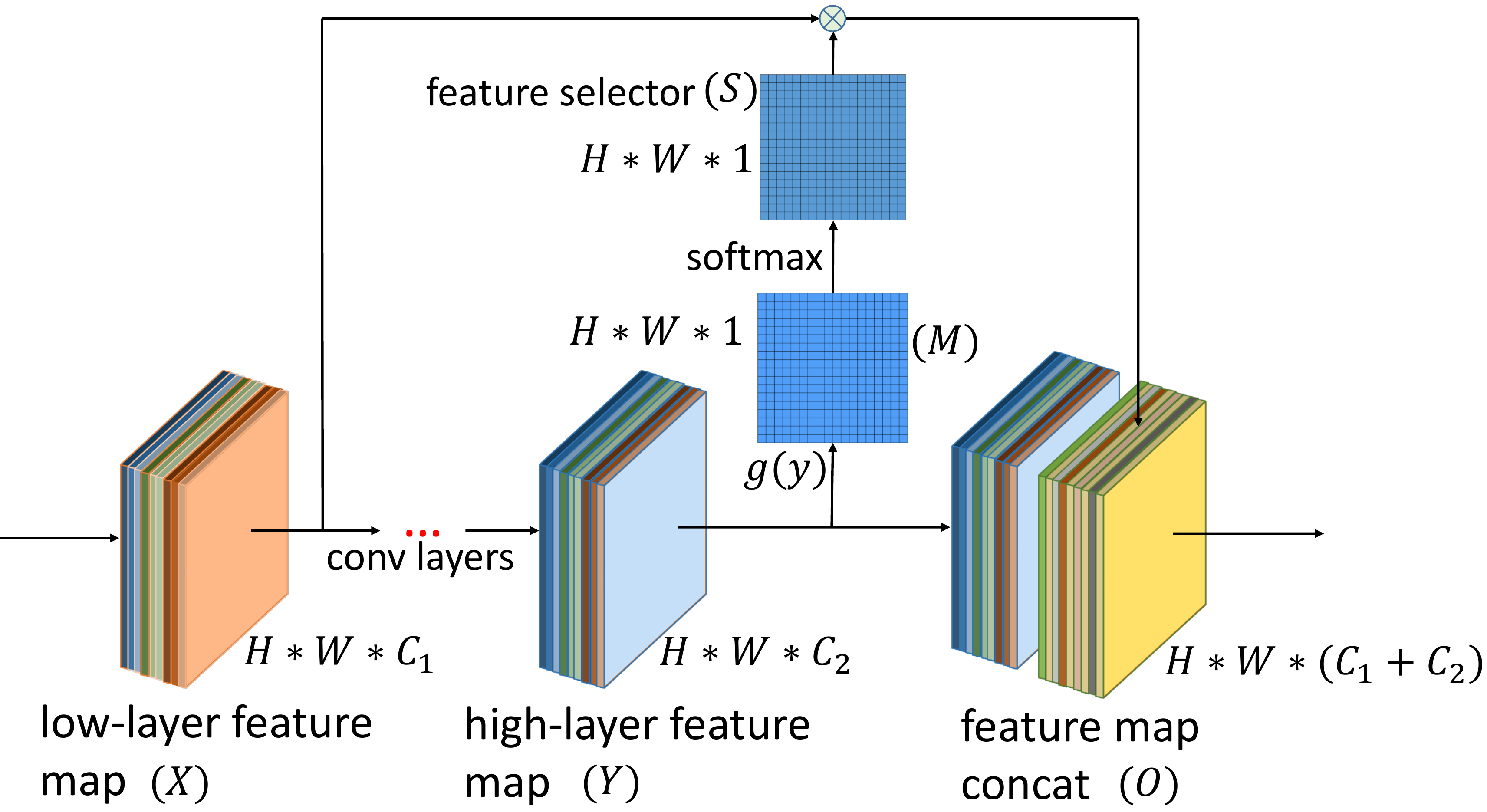}}
\hfil
\subfigure[]{\includegraphics[scale=0.25]{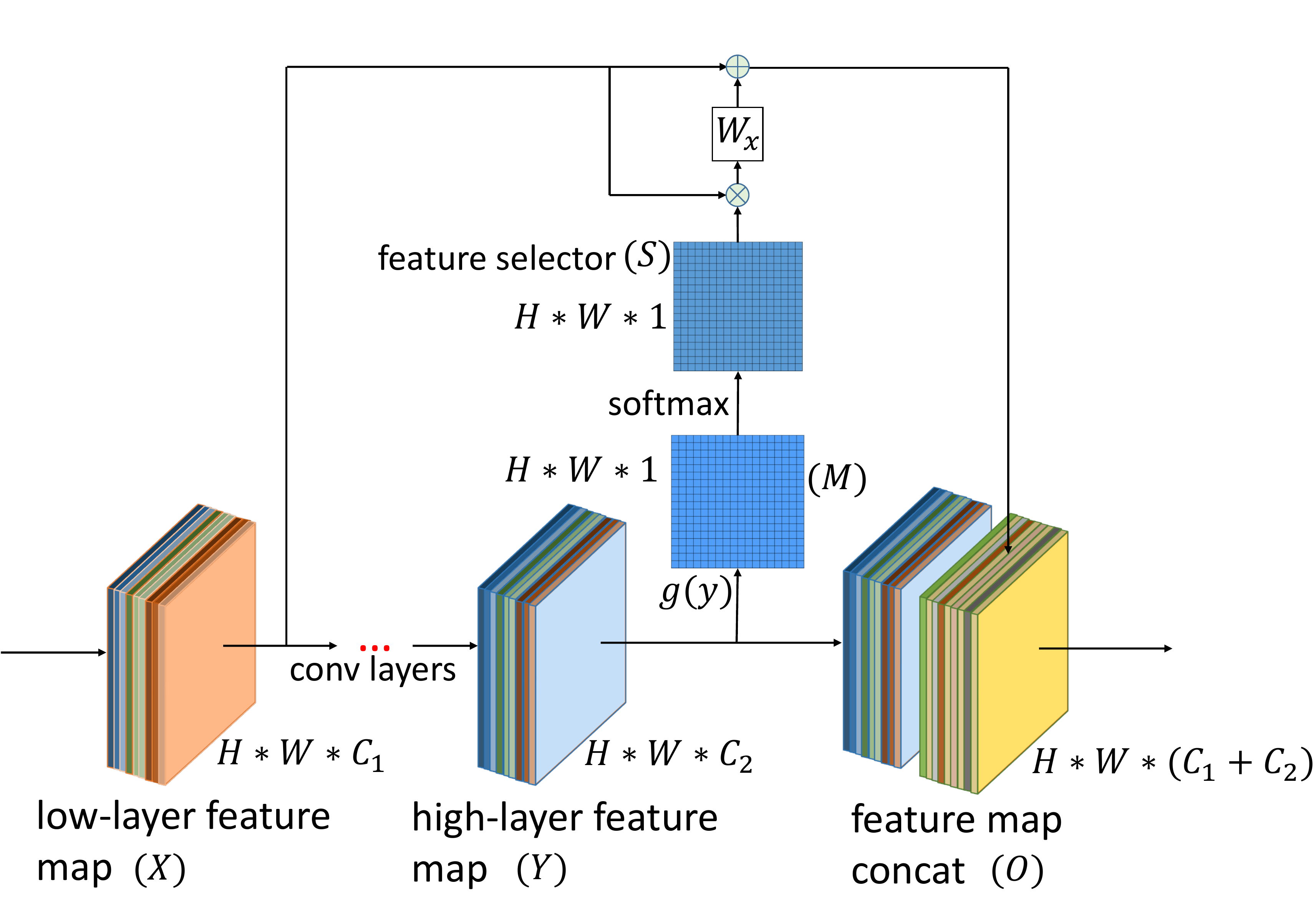}}
\caption{The proposed SFCM. "$\otimes$" denotes matrix multiplication, "$\oplus$" denotes element-wise sum. (a)Direct connection. (b)Residual connection.}
\label{Fig4}
\end{figure}

\section{Method}

\subsection{Selective Feature Connection Mechanism}
In the existing features concatenation methods,  feature maps obtained from different layers are combined directly. As shown in Fig.\ref{Fig3}, the combined features:
\begin{equation}
O = [X,Y]
\end{equation}
where $X\in \mathbf{R}^{H\times W\times C_1}$ denotes the low-layer feature map. $C_1$ is the number of channels, $W$ and $H$ denote the width and height. $Y\in \mathbf{R}^{H\times W\times C_2}$ is the high-level features of a ConvNet. $O\in \mathbf{R}^{H\times W\times (C_1+C_2)}$ is the combined features.

However, direct concatenation can not sufficiently apply the complementary information of high-layer and low-layer features. High-layer features can represent semantic information of an image and low-layer features contain more detail information. When directly combining the high-layer and low-layer features, the background clutter and semantic ambiguity may be caused due to the introduction of detailed information.

Inspired by the attention mechanism and recent advances in the deep neural network, we propose the SFCM. We assign attention scores for each local position on the low-layer feature map which indicates the importance of the low-layer features.
As shown in Fig.\ref{Fig4}, the attention scores $M$ are learned by $C_2\times 1\times 1\times 1$ convolutional filters.

\begin{equation}
M =  W_g * Y
\end{equation}
where "$*$" denotes a convolution operation. $W_g$ is the learned weight matrices, which is implemented as $1\times 1$ convolution.

In order to ensure the non-negative of the feature selector, we subsequently use the softmax normalization on $M$  to get feature selector $S$:
\begin{equation}
S_{i,j}= \frac{exp(M_{i,j})}{\sum_{i=1}^W\sum_{j=1}^H exp(M_{i,j})}
\end{equation}
In this equation, $S\in \mathbf{R}^{H\times W\times1}$, and $S_{i,j}$ is the score at position $(i,j)$.

The feature selector is learned to indicate the importance of the low-layer features. Thus,  more noteworthy features in lower-layer can be screened out by multiplying the feature selector with the low-layer features. We denote the value in $X$ corresponding to channel $c$ and spatial location $(i,j)$ as $X_{i,j,c}$. Thus, the new lower-layer features $X^s$ is given by
\begin{equation}
X^s_{i,j,c} = S_{i,j}*X_{i,j,c}
\end{equation}

Then the new lower-layer features are used to connect with higher-layer features.

\subsection{Adding SFCM To  Deep ConvNets }
We give two connection modes to add our method to a deep ConvNets, direct connection and residual connection \cite{he2016deep,wang2018non}. Fig.\ref{Fig4} shows the building block that constructs our SFCM.
\textbf{Direct connection:} In the direct connection, as shown in Fig.\ref{Fig4} (a),  $X^s$, given in Eq.(3), is concatenated to the higher level features $Y$:
\begin{equation}
O = [X^s,Y]
\end{equation}

\textbf{Residual connection:} As exemplify in Fig.\ref{Fig4} (b), the selective feature connection method can also be incorporated into many existing architectures in the form of residuals.

We further multiply $X^s$ by a scale parameter $W_x$ and add back the lower-layer features$X$:
\begin{equation}
X^\prime = W_x*X^s + X
\end{equation}
where $X^s$ is given in Eq.(3) and "$+X$" denotes a residual connection.

The residual connection allows us to insert the Selective Feature Connection block into any pre-trained model, without breaking its initial behavior.In the residual connection, $X^\prime$ is concatenated to the higher level features $Y$:
\begin{equation}
O = [X^\prime,Y]
\end{equation}

\subsection{Discussion}
We discuss why the SFCM is efficient by visualizing the feature layer of the network and the feature selector in SFCM, and then we discuss what exactly learned of the feature selector and how the SFCM works on improving state-of-art model performance. Fig.\ref{Fig2} shows an example of low-layer features selected for further use by a feature selector which is generated by high-level features. As shown in Fig.\ref{Fig2}, the "Feature selector" can learn to enhance the area of foreground and suppress the area of background. With the SFCM, most of the pixels in the low-level feature map are suppressed, so it can achieve better performance as it can combine high-resolution maps with low-level features without harm the semantic representation capacity of high-layer features and add more detail features of the foreground to further increase the representation ability of features.

\section{Experiments}
We perform experiments on a variety of tasks, like image classification, scene text detection, and image-to-image translation, to explore the effectiveness and universality of our SFCM. The output forms of these tasks, \emph{Label-Output} in classification, \emph{Coordinate-Regression} in detection and \emph{Image-Output} in image-to-image translation,  contain all output types in computer vision tasks, so the experiments can prove the generality of our method.

\begin{table*}[!t]
\begin{center}
\caption{Error rates(\%) on CIFAR datesets. "+" indicates standard data augmentation (translation and/or mirroring).The "Direct Connection" refers to use the proposed SFCM in the Direct Connection mode. The  error rates of direct and residual connections on CIFAR-10 and CIFAR-100 are both significantly lower than that of the DenseNet. Residual connection is more effective on existing architectures, as the selective feature connection method does not change the existing architectures in the form of residuals.}
\begin{tabular}{c|c|cc|cc|cc|cc}
\hline
\multicolumn{2}{c|}{\multirow{2}{*}{Method}} & \multicolumn{4}{c|}{Direct Connection} & \multicolumn{4}{|c}{Residual Connection} \\ \cline{3-10}
\multicolumn{2}{l|}{}                                         & C10     & C10+    & C100    & C100+    & C10      & C10+     & C100    & C100+    \\ \hline \hline
\multicolumn{2}{l|}{DenseNet(baseline)}                       & 7.00    & 5.24    & 27.55   & 24.42    & 7.00     & 5.24     & 27.55   & 24.42       \\ \hline
\multicolumn{2}{l|}{The first dense block with SFCM}          & 6.28    & 4.26    & 25.72   & 22.17    & 6.21     & 4.21     & 24.94   & 23.26    \\
\multicolumn{2}{l|}{The first two dense blocks with SFCM}     & 6.10    & 4.02    & 24.37   & 20.79    & 5.93     & 3.94     & 23.76   & 20.51    \\
\multicolumn{2}{l|}{All dense blocks with SFCM}               & \textbf{5.92}    & \textbf{3.81}    & \textbf{22.96}   & \textbf{19.31}    & \textbf{5.62}     & \textbf{3.73}     & \textbf{22.18}   & \textbf{19.27}    \\ \hline
\end{tabular}
\end{center}
\label{Tab1}
\end{table*}

\begin{table*}[]

\begin{center}
\caption{Results on ICDAR 2015 Challenge 4 Incidental Scene Text Localization task and COCO-Text. \textbf{*} This baseline is provided by the published model by EAST.}
\begin{tabular}{c|c|c|c|c|c|c}
\hline
\multicolumn{1}{c|}{\multirow{2}{*}{Method}} & \multicolumn{3}{c|}{ICDAR2015} & \multicolumn{3}{|c}{COCO-Text} \\ \cline{2-7}
\multicolumn{1}{c|}{}                        & Recall  & Precision  & F-score & Recall  & Precision  & F-score \\ \hline
EAST-ResNet(baseline)                        & 0.7722  & 0.8464     & 0.8076  & 0.3240  & 0.5039     & 0.3945  \\ \hline
EAST with SFCM                               & \textbf{0.7916}  & \textbf{0.8687}     & \textbf{0.8254}  & \textbf{0.3928}  & \textbf{0.5416}     & \textbf{0.4554}  \\ \hline
\end{tabular}
\end{center}
\label{Tab2}
\end{table*}

\subsection{Image classification:}
\textbf{Datasets:} The effect of SFCM in image classification is tested on the CIFAR-10 and CIFAR-100 datasets \cite{krizhevsky2009learning}. CIFAR-10 (C10) and CIFAR-100 (C100) consist of images in 10 classes and 100 classes respectively and all images in the two CIFAR datasets are colored natural images with 32¡Á32 pixels. The training sets contain 50,000 images and the test sets contain 10,000 images. In the training progress we hold out 5,000 training images as a validation set. A standard data augmentation scheme (mirroring/shifting)\cite{he2016deep,lin2013network} is  used for these two datasets  and  we denote this data augmentation scheme by a ¡°+¡± mark at the end of the dataset name (e.g., C10+).

\begin{figure}[!t]
\centering
\subfigure[]{\includegraphics[width=8.5cm]{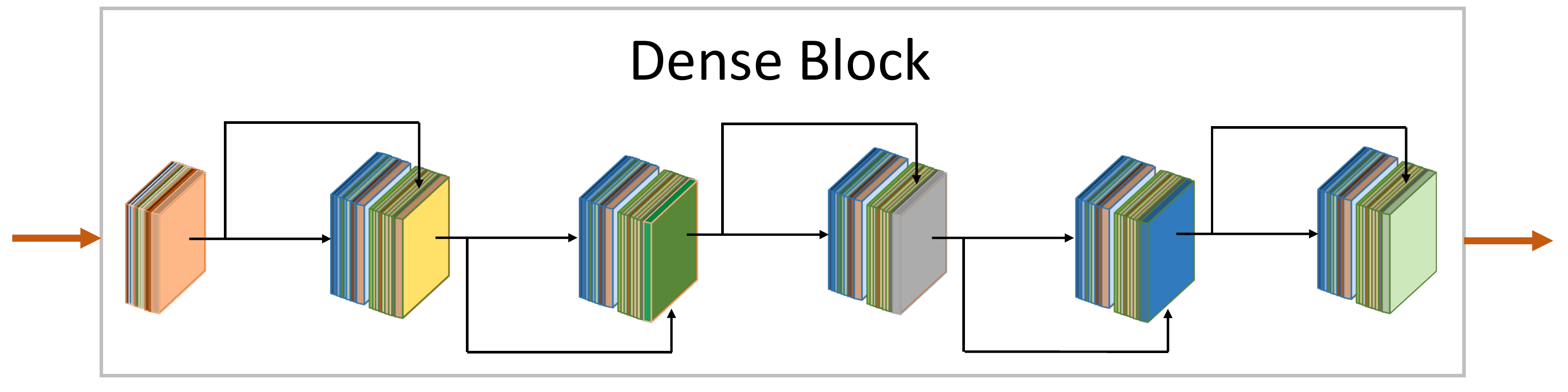}}
\hfil
\subfigure[]{\includegraphics[width=8.5cm]{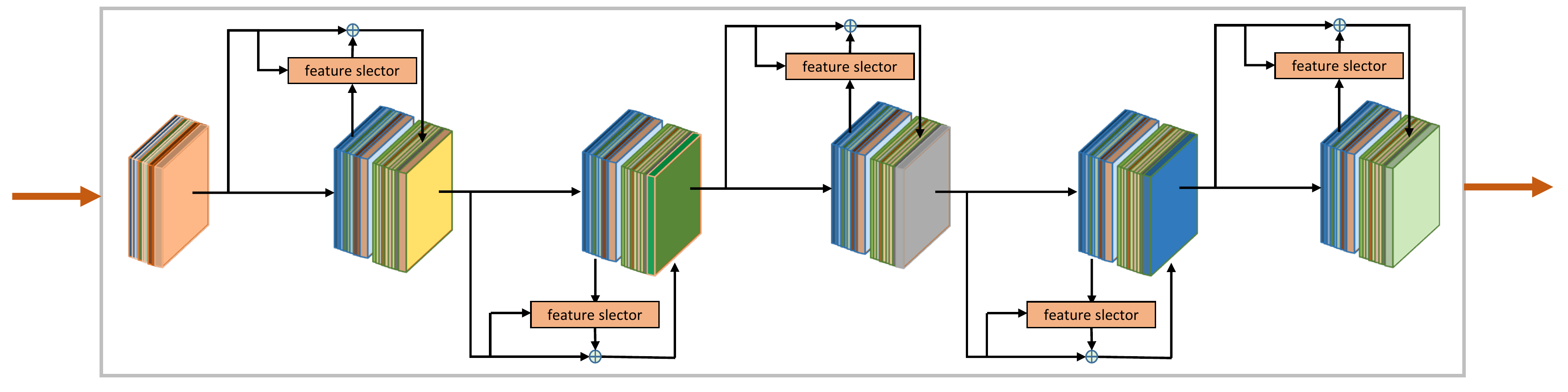}}
\caption{The dense block for image classification. (a) A dense block in DenseNet. (b) A dense block with SFCM (shown in orange).}
\label{Fig5}
\end{figure}

\textbf{Base Network:} We adopt the DenseNet \cite{huang2017densely} as the base model and change the dense block to include the SFCM. Fig.\ref{Fig5} (a) shows a dense block in DenseNet.A dense block with SFCM is shown in Fig.\ref{Fig5} (b).
Our focus is on the behaviors of a deep network with SFCM, but not on pushing the state-of-the-art results, so we intentionally use the simplest architectures in the DenseNet (three dense blocks, $k = 12$, $depth = 40$).

All the networks are trained end-to-end using stochastic gradient descent(SGD) and batch size 64 for 300 epochs. The initial learning rate is set to 0.1 and is divided by 10 at 50\% and 75\% of the total number of training epochs.

\textbf{Result Discussion:} Table\ref{Tab1} evaluates the effect of SFCM on DenseNet. When using the proposed SFCM in the first dense block, the error rates of 6.28\% with direct connection and 6.21\% with residual connection are both lower than that of 7.00\% achieved by the DenseNet without SFCM achieves on CIFAR-10 dataset. When using the proposed SFCM in the first two dense blocks, the error rates of 6.10\% with direct connection and 6.21\% with residual connection are both lower than that of 7.00\% achieved by the DenseNet without SFCM achieves on CIFAR-10 dataset. And the experimental results clearly show that:
\begin{itemize}
\item The accuracy is steadily improved when more layers use the proposed SFCM. When changing all the three dense blocks with the SFCM, the error drops to 5.62\% which is close to 20\% lower than the baseline.
\item Both the direct connection and residual connection significantly improve the accuracy of classification. The error rates of direct and residual connections on both CIFAR-10 and CIFAR-100 are both significantly lower than those of the DenseNet. This suggests that the SFCM can improve the representation learning ability of CNN models.
\item Residual connection is more effective on existing architectures, as the selective feature connection method does not change the existing architectures in the form of residuals.
\end{itemize}

\subsection{Scene text detection}
We further verify the effectiveness of our method in scene text detection. Scene text detection is challenging due to text may exist in natural images with arbitrary size and orientation. The core of text detection is the design of features to distinguish text from backgrounds. Merging feature maps of different layers should be possible to improve the performance of detecting text in various size.

\textbf{Datasets:} We use the  ICDAR 2015 Incidental Text datasets \cite{karatzas2015icdar} and the COCO-Text \cite{veit2016coco} to  test the effect of SFCM on Scene text detection.
The ICDAR 2015 Incidental Text datasets issues from Challenge 4 of the ICDAR 2015 Robust Reading Competition. It includes 1000 training images and 500 testing images. These images are taken by Google Glass in an incidental way. Therefore text in the scene can be in arbitrary orientations, or suffer from motion blur and low resolution. Each image may contain multi-oriented text. Annotations are provided in terms of word bounding boxes.
COCO-Text is a large dataset which contains 63686 images, where 43,686 of the images is used for training, 10,000 for validation, and 10,000 for testing. It is one of the challenges of ICDAR 2017 Robust Reading Competition.

\begin{figure}[!t]
\centering
\includegraphics[height=7.5cm]{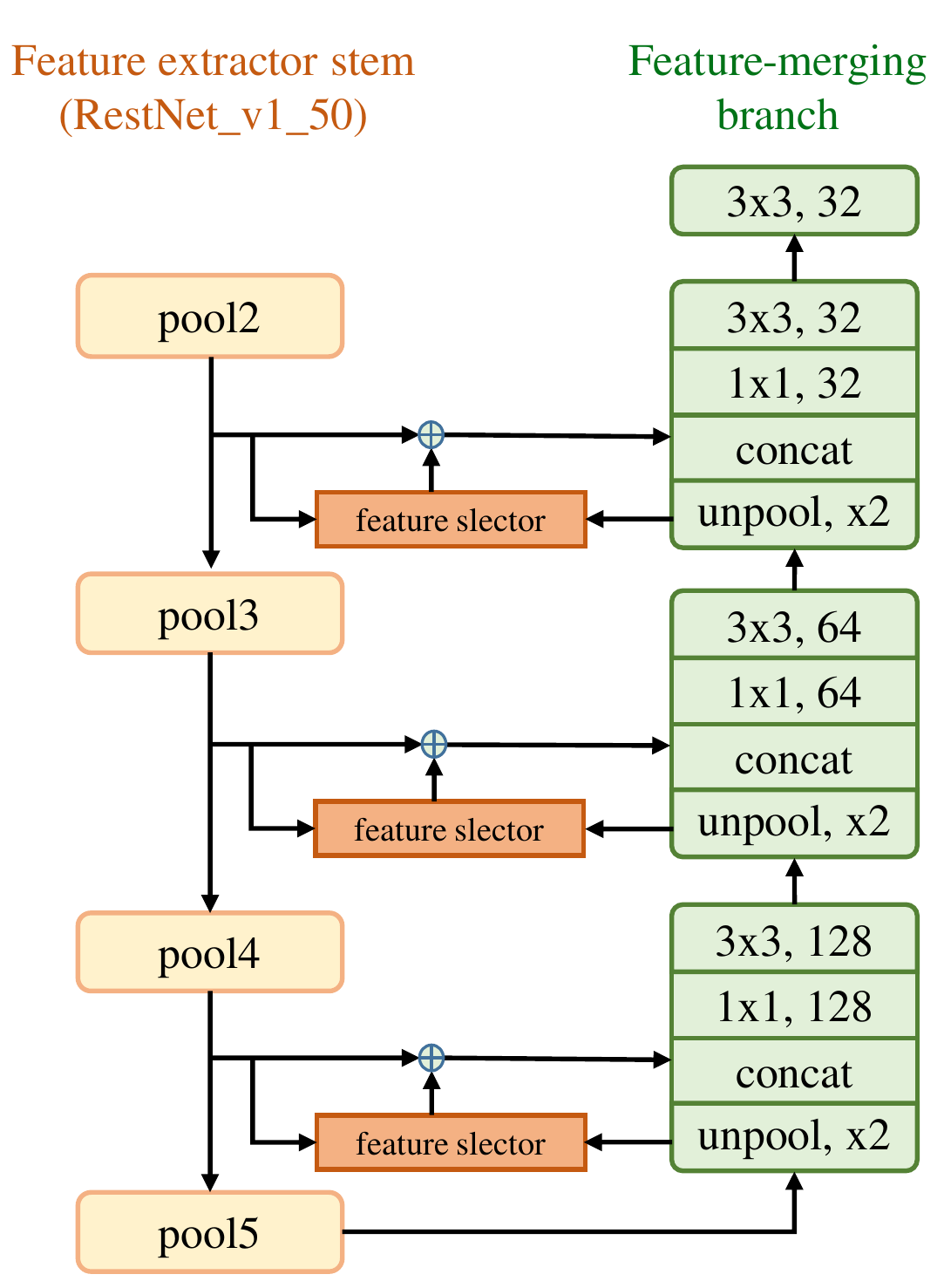}
\caption{The feature-merging branch of the EAST is modified with SFCM.}
\label{Fig6}
\end{figure}

\textbf{Base Network:} We use the EAST \cite{zhou2017east} as the base model and only change the feature-merging branch with SFCM(Fig.\ref{Fig6}). We take the residual connection of the SFCM because it is more effective on existing architectures. The network is trained end-to-end using ADAM \cite{kingma2014adam} optimizer. To speed up learning, we uniformly sample 512x512 crops from images to form a minibatch of size 14 on one GPU. Learning rate of ADAM starts from 1e-3 and is decayed to its 0.94 every 10000 minibatches. The network is trained until performance stops improving.

\begin{figure}[!t]
\centering
\subfigure[]{\includegraphics[width=8cm]{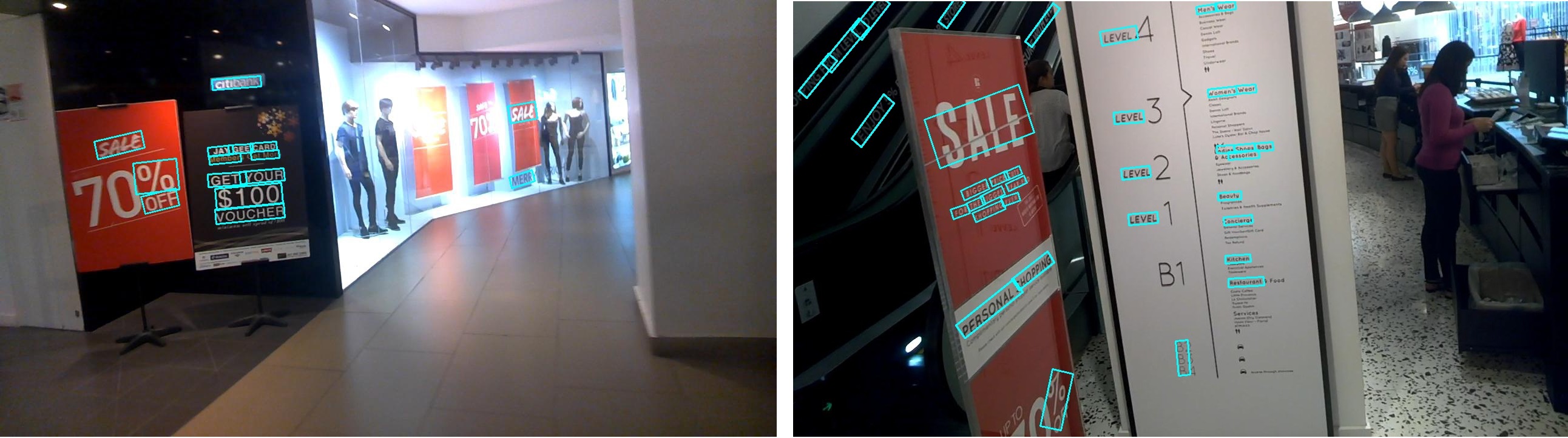}}
\hfil
\subfigure[]{\includegraphics[width=8cm]{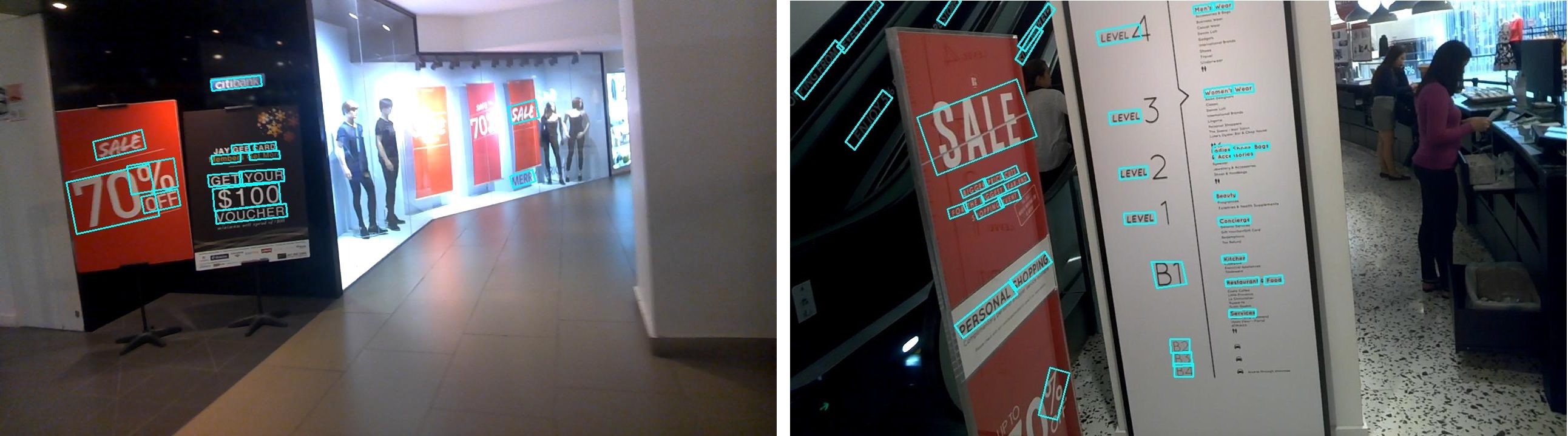}}
\caption{Some results on images in ICDAR-15 dataset. (a)Result of EAST-ResNet(baseline). (b)Result of EAST-ResNet with SFCM.}
\label{Fig7}
\end{figure}

\textbf{Result Discussion:} Table \ref{Tab2} evaluates the effect of SFCM on EAST. When changing the feature-merging branch with the SFCM, the EAST achieves an F-score of 0.8254 on ICDAR 2015 dataset and 0.4554 on COCO-Text dataset, and both accuracy and recall rate have been greatly improved on these two datasets. This is because the model with the proposed connection mechanism reduces the misjudgment of small targets and makes the detection of larger targets more accurate, as shown in Fig.\ref{Fig7}.  This further verifies that our SFCM can concatenate the low-level features and high-level features more effectively.

\begin{table}[]\small
\begin{center}
\caption{Segmentation accuracy for the Cityscapes labels$\leftrightarrow$photos task. The cGAN with SFCM outperform the original cGAN \cite{isola2017image}.}
\begin{tabular}{l|c|c|c}
\hline
Method          & Per-pixel acc. & Per-class acc. & Class IOU \\ \hline \hline
cGAN (baseline) & 0.66           & 0.23           & 0.17      \\ \hline
cGAN with SFCM  & \textbf{0.71}          & \textbf{ 0.24}           & 0.17      \\ \hline \hline
Ground truth    & 0.80           & 0.26           & 0.21      \\ \hline
\end{tabular}
\end{center}

\label{Tab3}
\end{table}

\begin{figure}[!t]
\centering
\subfigure[]{\includegraphics[width=8cm]{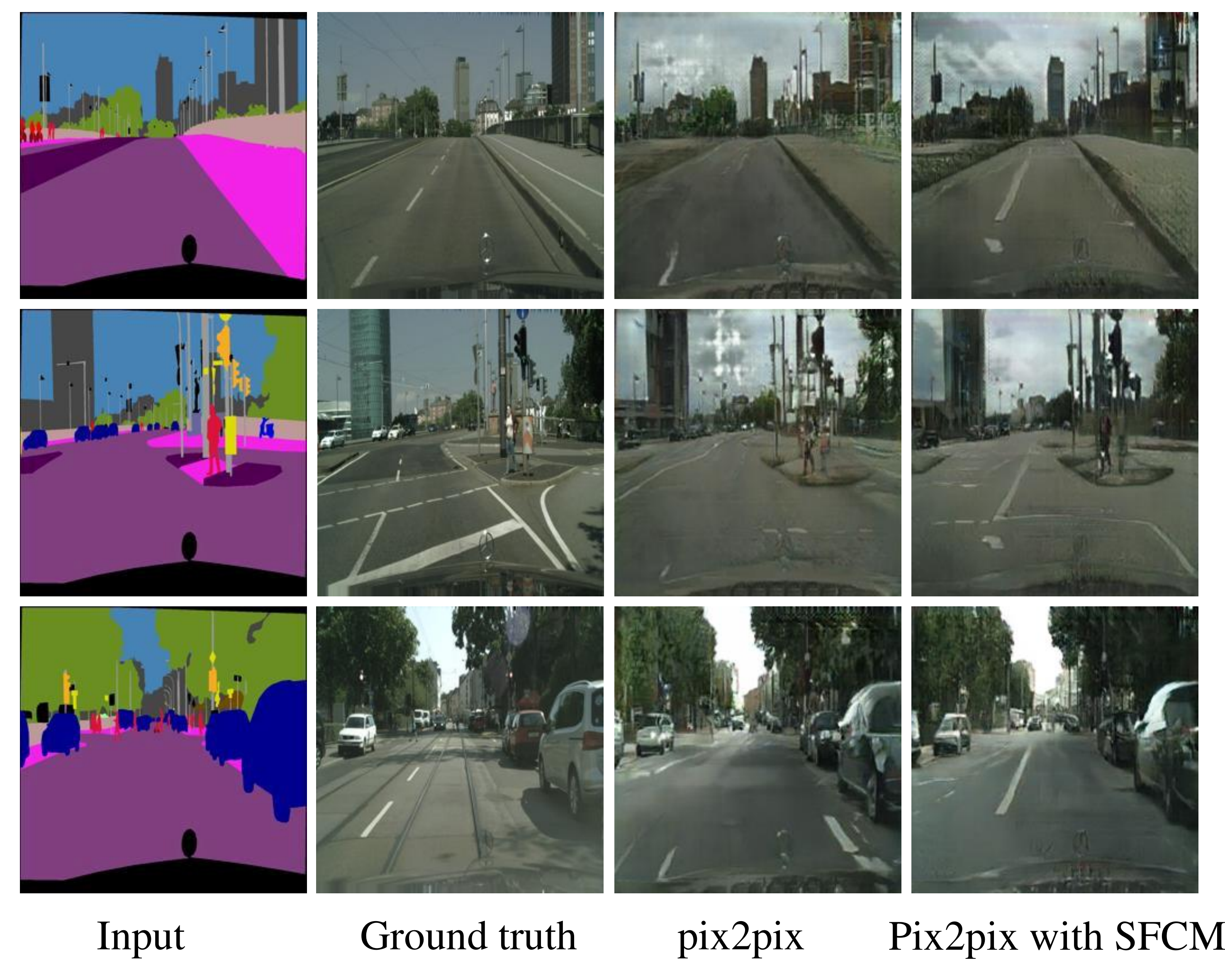}}
\hfil
\subfigure[]{\includegraphics[width=8cm]{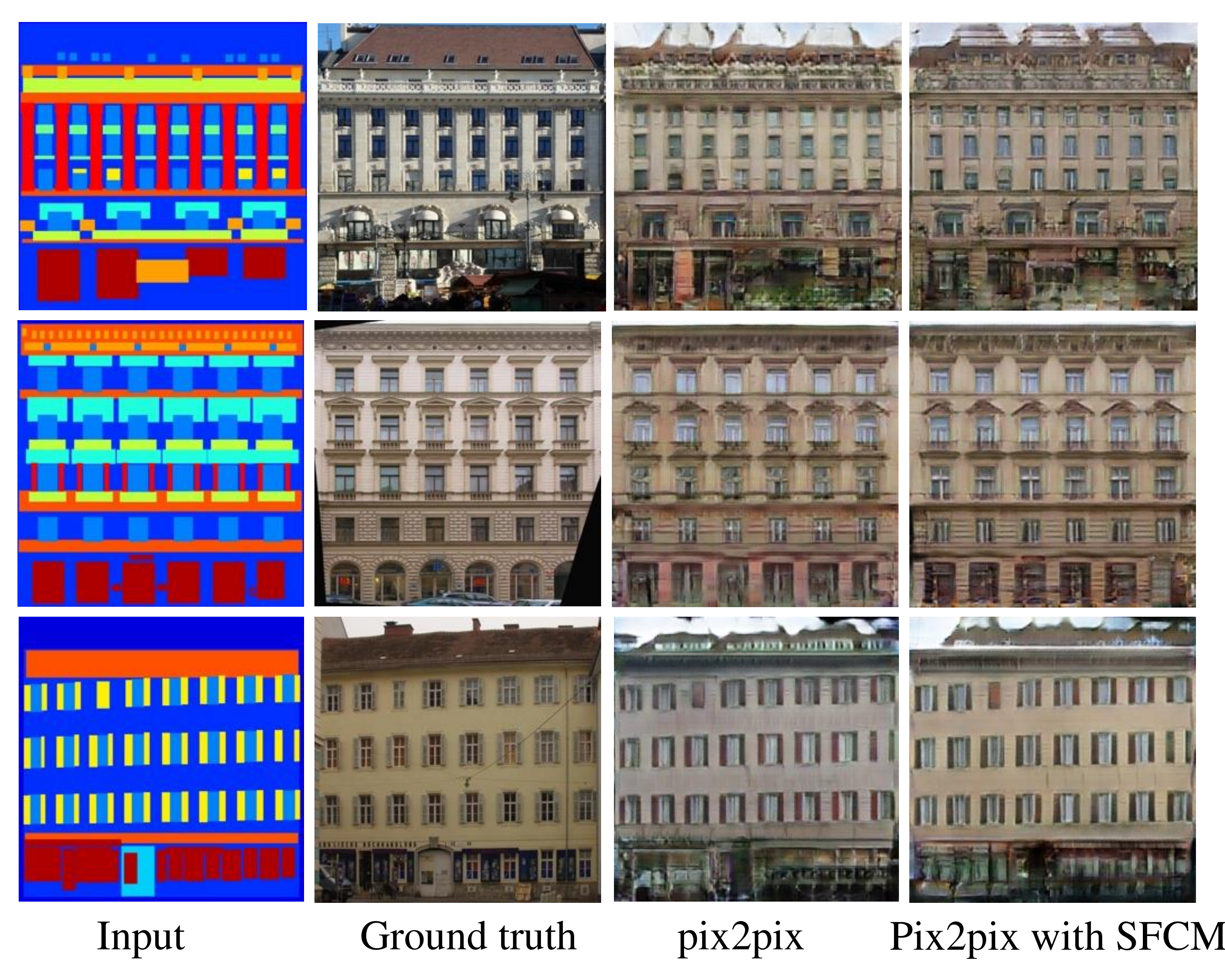}}
\caption{Some results on the image-to-image translation tasks. The pix2pix model with the SFCM results in much higher quality results, especially in details. (a) Example results on the Cityscapes dataset. (b) Example results on CMP Facades.}
\label{Fig8}
\end{figure}

\subsection{Image-to-image translation}
Image-to-image translation is one of the image generation tasks aiming to translate an input image into a corresponding output image. A defining feature of image-to-image translation problems is that they map a high-resolution input grid to a high-resolution output grid. Combining high-level semantic information with low-level detail information effectively can significantly improve the performance of the translation tasks. We use the pix2pix (cGAN) \cite{isola2017image} as the base model to test the effect of SFCM. This model solves the image generation task that the input and output differ in surface appearance, but both are renderings of the same underlying structure. We test the method on two tasks and datasets:
\begin{itemize}
\item[1.] Semantic labels to photo, trained on the Cityscapes dataset \cite{cordts2016cityscapes}
\item[2.] Architectural labels to photo, trained on CMP Facades \cite{tylevcek2013spatial}
\end{itemize}

Quantitative evaluation of generative models is known to be challenging. In this paper, we use the FCN-score \cite{isola2017image} to evaluate the performance of our SFCM. We adopt the popular FCN-8s \cite{long2015fully} architecture for semantic segmentation, and train it on the cityscapes dataset. We then score synthesized photos by the classification accuracy against the labels these photos were synthesized from.

Table \ref{Tab3} evaluates the effect of SFCM on cGAN. The cGAN with SFCM outperforms the original cGAN.Some results are shown in Fig.\ref{Fig8}. The resolution of the images translated by the pix2pix is low and the image edges are illegible. With the SFCM, images translated by the pix2pix show much higher quality especially in details. This suggests that the SFCM has learned a more effective way to combine high-level semantic information with low-level detail information in the training of a neural network model.

\section{Conclusion and Future Work}
We have presented the SFCM, a general network architecture to concatenate CNN features of different layers, and given two connection modes, direct connection and residual connection, to use this architecture.
With our SFCM, low-level features are selectively linked to high-level features with an attention map which is generated by high-level features.  Experiments show that our method is effective and generic on a variety of tasks, including image classification, scene text detection, and image-to-image translation. Thus, it provides a practical solution for research and applications to use multi-layer CNN Features more effectively.

{\small
\bibliographystyle{ieee}
\bibliography{egbib}

\begin{thebibliography}{10}\itemsep=-1pt

\bibitem{bell2016inside}
S.~Bell, C.~Lawrence~Zitnick, K.~Bala, and R.~Girshick.
\newblock Inside-outside net: Detecting objects in context with skip pooling
  and recurrent neural networks.
\newblock In {\em Proceedings of the IEEE conference on computer vision and
  pattern recognition}, pages 2874--2883, 2016.

\bibitem{cai2016unified}
Z.~Cai, Q.~Fan, R.~S. Feris, and N.~Vasconcelos.
\newblock A unified multi-scale deep convolutional neural network for fast
  object detection.
\newblock In {\em European Conference on Computer Vision}, pages 354--370.
  Springer, 2016.

\bibitem{chen2016attention}
L.-C. Chen, Y.~Yang, J.~Wang, W.~Xu, and A.~L. Yuille.
\newblock Attention to scale: Scale-aware semantic image segmentation.
\newblock In {\em Proceedings of the IEEE conference on computer vision and
  pattern recognition}, pages 3640--3649, 2016.

\bibitem{cordts2016cityscapes}
M.~Cordts, M.~Omran, S.~Ramos, T.~Rehfeld, M.~Enzweiler, R.~Benenson,
  U.~Franke, S.~Roth, and B.~Schiele.
\newblock The cityscapes dataset for semantic urban scene understanding.
\newblock In {\em Proceedings of the IEEE conference on computer vision and
  pattern recognition}, pages 3213--3223, 2016.

\bibitem{dai2016r}
J.~Dai, Y.~Li, K.~He, and J.~Sun.
\newblock R-fcn: Object detection via region-based fully convolutional
  networks.
\newblock In {\em Advances in neural information processing systems}, pages
  379--387, 2016.

\bibitem{dai2017deformable}
J.~Dai, H.~Qi, Y.~Xiong, Y.~Li, G.~Zhang, H.~Hu, and Y.~Wei.
\newblock Deformable convolutional networks.
\newblock {\em CoRR, abs/1703.06211}, 1(2):3, 2017.

\bibitem{girshick2015fast}
R.~Girshick.
\newblock Fast r-cnn.
\newblock In {\em Proceedings of the IEEE international conference on computer
  vision}, pages 1440--1448, 2015.

\bibitem{gregor2015draw}
K.~Gregor, I.~Danihelka, A.~Graves, D.~J. Rezende, and D.~Wierstra.
\newblock Draw: A recurrent neural network for image generation.
\newblock {\em arXiv preprint arXiv:1502.04623}, 2015.

\bibitem{he2017mask}
K.~He, G.~Gkioxari, P.~Doll{\'a}r, and R.~Girshick.
\newblock Mask r-cnn.
\newblock In {\em Computer Vision (ICCV), 2017 IEEE International Conference
  on}, pages 2980--2988. IEEE, 2017.

\bibitem{he2016deep}
K.~He, X.~Zhang, S.~Ren, and J.~Sun.
\newblock Deep residual learning for image recognition.
\newblock In {\em Proceedings of the IEEE conference on computer vision and
  pattern recognition}, pages 770--778, 2016.

\bibitem{huang2017densely}
G.~Huang, Z.~Liu, L.~Van Der~Maaten, and K.~Q. Weinberger.
\newblock Densely connected convolutional networks.
\newblock In {\em CVPR}, volume~1, page~3, 2017.

\bibitem{isola2017image}
P.~Isola, J.-Y. Zhu, T.~Zhou, and A.~A. Efros.
\newblock Image-to-image translation with conditional adversarial networks.
\newblock {\em arXiv preprint}, 2017.

\bibitem{jaderberg2015spatial}
M.~Jaderberg, K.~Simonyan, A.~Zisserman, et~al.
\newblock Spatial transformer networks.
\newblock In {\em Advances in neural information processing systems}, pages
  2017--2025, 2015.

\bibitem{karatzas2015icdar}
D.~Karatzas, L.~Gomez-Bigorda, A.~Nicolaou, S.~Ghosh, A.~Bagdanov, M.~Iwamura,
  J.~Matas, L.~Neumann, V.~R. Chandrasekhar, S.~Lu, et~al.
\newblock Icdar 2015 competition on robust reading.
\newblock In {\em Document Analysis and Recognition (ICDAR), 2015 13th
  International Conference on}, pages 1156--1160. IEEE, 2015.

\bibitem{kingma2014adam}
D.~P. Kingma and J.~Ba.
\newblock Adam: A method for stochastic optimization.
\newblock {\em arXiv preprint arXiv:1412.6980}, 2014.

\bibitem{kong2016hypernet}
T.~Kong, A.~Yao, Y.~Chen, and F.~Sun.
\newblock Hypernet: Towards accurate region proposal generation and joint
  object detection.
\newblock In {\em Proceedings of the IEEE conference on computer vision and
  pattern recognition}, pages 845--853, 2016.

\bibitem{krizhevsky2009learning}
A.~Krizhevsky and G.~Hinton.
\newblock Learning multiple layers of features from tiny images.
\newblock Technical report, Citeseer, 2009.

\bibitem{krizhevsky2012imagenet}
A.~Krizhevsky, I.~Sutskever, and G.~E. Hinton.
\newblock Imagenet classification with deep convolutional neural networks.
\newblock In {\em Advances in neural information processing systems}, pages
  1097--1105, 2012.

\bibitem{lin2013network}
M.~Lin, Q.~Chen, and S.~Yan.
\newblock Network in network.
\newblock {\em arXiv preprint arXiv:1312.4400}, 2013.

\bibitem{lin2017feature}
T.-Y. Lin, P.~Doll{\'a}r, R.~B. Girshick, K.~He, B.~Hariharan, and S.~J.
  Belongie.
\newblock Feature pyramid networks for object detection.
\newblock In {\em CVPR}, volume~1, page~4, 2017.

\bibitem{liu2016ssd}
W.~Liu, D.~Anguelov, D.~Erhan, C.~Szegedy, S.~Reed, C.-Y. Fu, and A.~C. Berg.
\newblock Ssd: Single shot multibox detector.
\newblock In {\em European conference on computer vision}, pages 21--37.
  Springer, 2016.

\bibitem{liu2015parsenet}
W.~Liu, A.~Rabinovich, and A.~C. Berg.
\newblock Parsenet: Looking wider to see better.
\newblock {\em arXiv preprint arXiv:1506.04579}, 2015.

\bibitem{long2015fully}
J.~Long, E.~Shelhamer, and T.~Darrell.
\newblock Fully convolutional networks for semantic segmentation.
\newblock In {\em Proceedings of the IEEE conference on computer vision and
  pattern recognition}, pages 3431--3440, 2015.

\bibitem{redmon2016you}
J.~Redmon, S.~Divvala, R.~Girshick, and A.~Farhadi.
\newblock You only look once: Unified, real-time object detection.
\newblock In {\em Proceedings of the IEEE conference on computer vision and
  pattern recognition}, pages 779--788, 2016.

\bibitem{ren2015faster}
S.~Ren, K.~He, R.~Girshick, and J.~Sun.
\newblock Faster r-cnn: Towards real-time object detection with region proposal
  networks.
\newblock In {\em Advances in neural information processing systems}, pages
  91--99, 2015.

\bibitem{ronneberger2015u}
O.~Ronneberger, P.~Fischer, and T.~Brox.
\newblock U-net: Convolutional networks for biomedical image segmentation.
\newblock In {\em International Conference on Medical image computing and
  computer-assisted intervention}, pages 234--241. Springer, 2015.

\bibitem{tylevcek2013spatial}
R.~Tyle{\v{c}}ek and R.~{\v{S}}{\'a}ra.
\newblock Spatial pattern templates for recognition of objects with regular
  structure.
\newblock In {\em German Conference on Pattern Recognition}, pages 364--374.
  Springer, 2013.

\bibitem{vaswani2017attention}
A.~Vaswani, N.~Shazeer, N.~Parmar, J.~Uszkoreit, L.~Jones, A.~N. Gomez,
  {\L}.~Kaiser, and I.~Polosukhin.
\newblock Attention is all you need.
\newblock In {\em Advances in Neural Information Processing Systems}, pages
  5998--6008, 2017.

\bibitem{veit2016coco}
A.~Veit, T.~Matera, L.~Neumann, J.~Matas, and S.~Belongie.
\newblock Coco-text: Dataset and benchmark for text detection and recognition
  in natural images.
\newblock {\em arXiv preprint arXiv:1601.07140}, 2016.

\bibitem{wang2017residual}
F.~Wang, M.~Jiang, C.~Qian, S.~Yang, C.~Li, H.~Zhang, X.~Wang, and X.~Tang.
\newblock Residual attention network for image classification.
\newblock {\em arXiv preprint arXiv:1704.06904}, 2017.

\bibitem{wang2016deeply}
J.~Wang, Z.~Wei, T.~Zhang, and W.~Zeng.
\newblock Deeply-fused nets.
\newblock {\em arXiv preprint arXiv:1605.07716}, 2016.

\bibitem{wang2018non}
X.~Wang, R.~Girshick, A.~Gupta, and K.~He.
\newblock Non-local neural networks.
\newblock In {\em The IEEE Conference on Computer Vision and Pattern
  Recognition (CVPR)}, 2018.

\bibitem{xu2015show}
K.~Xu, J.~Ba, R.~Kiros, K.~Cho, A.~Courville, R.~Salakhudinov, R.~Zemel, and
  Y.~Bengio.
\newblock Show, attend and tell: Neural image caption generation with visual
  attention.
\newblock In {\em International conference on machine learning}, pages
  2048--2057, 2015.

\bibitem{yang2016stacked}
Z.~Yang, X.~He, J.~Gao, L.~Deng, and A.~Smola.
\newblock Stacked attention networks for image question answering.
\newblock In {\em Proceedings of the IEEE Conference on Computer Vision and
  Pattern Recognition}, pages 21--29, 2016.

\bibitem{zeiler2014visualizing}
M.~D. Zeiler and R.~Fergus.
\newblock Visualizing and understanding convolutional networks.
\newblock In {\em European conference on computer vision}, pages 818--833.
  Springer, 2014.

\bibitem{zhou2017east}
X.~Zhou, C.~Yao, H.~Wen, Y.~Wang, S.~Zhou, W.~He, and J.~Liang.
\newblock East: an efficient and accurate scene text detector.
\newblock In {\em Proc. CVPR}, pages 2642--2651, 2017.

\end{thebibliography}
}

\end{document}